\documentclass[letterpaper, 10 pt, conference]{ieeeconf}  % Comment this line out if you need a4paper

\IEEEoverridecommandlockouts                              % This command is only needed if 
\usepackage{pdfpages}
\usepackage{float}
\usepackage{hyperref}
\usepackage{graphics} % for pdf, bitmapped graphics files
\title{\LARGE \bf
IMMERTWIN: A Mixed Reality Framework for Enhanced Robotic Arm Teleoperation
}

\author{Florent P Audonnet$^{1}$, Ixchel G. Ramirez-Alpizar$^{2}$ and Gerardo Aragon-Camarasa$^{1}$% $<$-this % stops a space
\thanks{This research has been supported by EPSRC DTA No. 2605103 and EPSRC Grant No. EP/S019472/1}% $<$-this % stops a space
\thanks{$^{1}$ School of Computing Science, University of Glasgow, G12 8QQ, Scotland, United Kingdom {\tt\small f.audonnet.1@research.gla.ac.uk; gerardo.aragoncamarasa@glasgow.ac.uk}}%
\thanks{$^{2}$ ICPS Research Center, National Institute of Advanced Industrial Science and Technology (AIST), Tokyo, Japan}
}

\begin{document}
\let\oldUrl\url
\renewcommand{\url}[1]{\href{#1}{Link}}

\maketitle
\thispagestyle{empty}
\pagestyle{empty}

%%%%%%%%%%%%%%%%%%%%%%%%%%%%%%%%%%%%%%%%%%%%%%%%%%%%%%%%%%%%%%%%%%%%%%%%%%%%%%%%
\begin{abstract}
 We present IMMERTWIN, a mixed reality framework for enhance robotic arm teleoperation using a closed-loop digital twin as a bridge for interaction between the user and the robotic system. We evaluated IMMERTWIN by performing a medium-scale user survey with 26 participants on two robots. Users were asked to teleoperate with both robots inside the virtual environment to pick and place 3 cubes in a tower and to repeat this task as many times as possible in 10 minutes, with only 5 minutes of training beforehand. Our experimental results show that most users were able to succeed by building at least a tower of 3 cubes regardless of the robot used and a maximum of 10 towers (1 tower per minute). In addition, users preferred to use IMMERTWIN over our previous work, TELESIM, as it caused them less mental workload. The project website and source code can be found at: \href{https://cvas-ug.github.io/immertwin}{https://cvas-ug.github.io/immertwin}
\end{abstract}

%%%%%%%%%%%%%%%%%%%%%%%%%%%%%%%%%%%%%%%%%%%%%%%%%%%%%%%%%%%%%%%%%%%%%%%%%%%%%%%%
\section{Introduction}\label{sec:intro}
The ANA Avatar XPRIZE \cite{noauthor_xprize_2023} competition has significantly increased interest in telepresence robotics. Telepresence in the competition refers to robots that can be fully controlled by a human operator from a remote location, thereby simulating the operator's presence at the robot's site. The need for a human operator persists due to the limitations of fully autonomous systems, which remain highly constrained and ineffective beyond predefined scenarios \cite{fu_learning_2024, sun_fully_2022}. Consequently, researchers have focused on integrating autonomous systems with direct teleoperation to enhance performance and alleviate the cognitive load on users. This integration often involves augmenting visual information or automating specific robot movements \cite{moniruzzaman_teleoperation_2022}. 

Despite recent advancements, the user's field of view remains restricted to the camera's perspective mounted on the robot in telepresence scenarios or to the user's immediate surroundings in direct teleoperation \cite{fu_learning_2024, audonnet_telesim_2023}. This limitation was evident in our previous work, TELESIM \cite{audonnet_telesim_2023}, where user mobility was restricted due to the robot's movement being directly linked to the user's hand movements in a one-to-one mapping. In this paper, we introduce IMMERTWIN, an immersive and modular plug-and-play framework for robotic arm teleoperation that extends the capabilities of TELESIM. IMMERTWIN situates the user within a Digital Twin environment, allowing for free movement and varied viewpoints while maintaining control over the robot.

\begin{figure}[!t]
    \centering
    \includegraphics[width=0.9\linewidth]{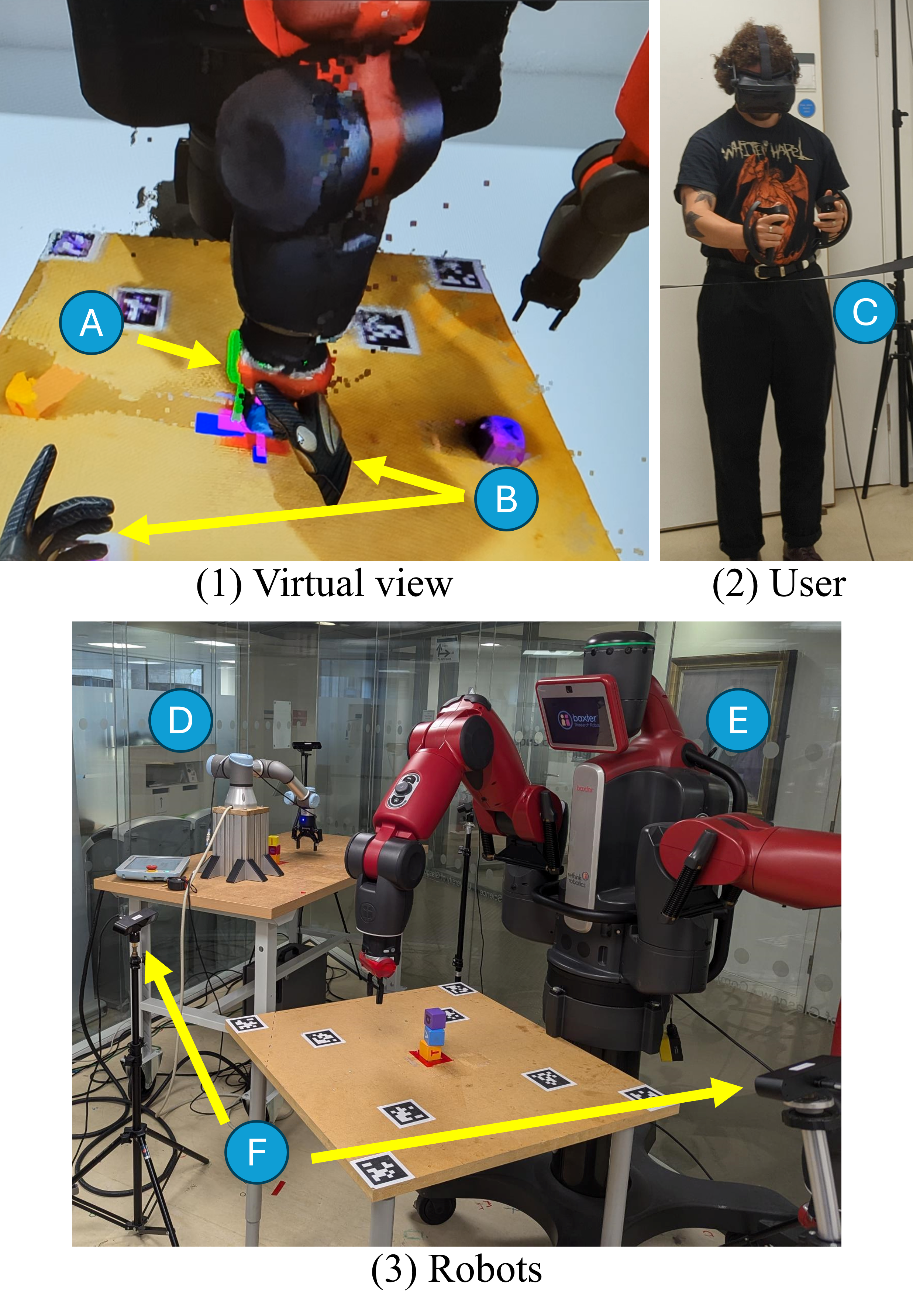}
    \caption{Our experimental setup comprises the following components. (1) The view from the user inside Unreal Engine, with their virtual hands (A) and the virtual gripper (B). (2) The user wearing the VR headset, with a black security tape (C) to avoid users walk towards the robots. (3) The room containing both the UR3 robot (D) and the Baxter robot (E), along with 4 ZED 2I cameras (F) to live stream a pointcloud into the virtual environment.}
    \label{fig:experimental_setup}
\end{figure}

We evaluate the performance and user-friendliness of IMMERTWIN through a comprehensive user study involving 26 participants. Each participant was tasked with controlling two robots to repeatedly perform a tower stacking task within ten minutes after a brief five-minute training session. Although none of the participants had prior experience with IMMERTWIN, some were familiar with TELESIM. Our results include cognitive and physical loads and we explore the influence of different robot types and prior experience on teleoperation performance. Our findings are then compared with those from TELESIM. Our contributions are:
\begin{itemize}
% \item A modular and plug-and-play framework for teleoperation applicable to any robotic arm using a digital twin.
\item Implementing an immersive mixed reality teleoperation system utilising near-real-time point cloud visualisation.
\item Experimental validation of the framework's performance using a straightforward task suitable for non-experts.
\item A rigorous evaluation involving 26 participants highlighting the user-friendliness of IMMERTWIN.
\end{itemize}

\section{Background}

Teleoperation allows remote control of robots and has been crucial in enabling Industry 5.0, emphasising human-robot synergy\cite{zhang_research_2023}. This technology effectively facilitates the training of robots to execute tasks autonomously, employing methods such as Imitation Learning and Reinforcement Learning \cite{fu_learning_2024, si_review_2021, he_learning_2024}. However, teleoperation can exert significant physical and mental stress on users \cite{audonnet_telesim_2023}. Moreover, achieving movements similar to those performed by humans during teleoperation remains challenging due to its dependence on factors, including the robot type, controller and the operator's prior experience \cite{audonnet_telesim_2023}. 

Researchers have investigated various types of controllers, including Virtual Reality (VR) \cite{wang_intent_2021, dafarra_icub3_2022}, joysticks \cite{aronson_eye-hand_2018}, haptic interfaces \cite{li_bilateral_2024, scherzinger_learning_2023}, and vision-based systems \cite{rosen_communicating_2019}. However, direct comparisons of their performance are rare, as studies often focus on enhancing the performance of a single robot or task using assistive methods \cite{moniruzzaman_teleoperation_2022}. These evaluations are typically conducted with small-scale user surveys involving fewer than 20 participants on a single robot\cite{lin_shared_2020, gottardi_shared_2022}.

The use of VR headsets to provide additional information through mixed reality has been a significant assistive improvement in teleoperation\cite{naughton_integrating_2024}. VR was initially used as a flight simulator in teleoperation \cite{basu_brief_2019} and has since been utilized to control robots. It played a pivotal role in the ANA Avatar XPRIZE challenge, where users teleoperate robots in remote locations, adapting to the robot's capabilities \cite{schwarz_robust_2023, dafarra_icub3_2022}. Typically, users in VR systems view only the camera feed (a 2D video stream) from the robot, which is common in most immersive VR teleoperation systems. 

Mixed Reality in robotic teleoperation blurs the line between real and digital environments by overlaying digital information on video streams or integrating real-world data into simulated environments. This technology is regarded as the next step in enhancing VR teleoperation \cite{su_integrating_2023}. Most research focuses on the former, such as \cite{naughton_integrating_2024}, where surface highlights are added for automatic robot alignment. 

Research indicates that visualising a real-time point cloud within a virtual environment yields better results. For example, Su \textit{et al.} \cite{su_mixed_2022} conducted a study with 15 participants to evaluate the cognitive and physical load of three systems: one rendering only 2D images, another using a stereoscopic system to supplement the 2D images, and a third rendering a 3D point-cloud. The point-cloud system demonstrated higher performance with reduced cognitive and physical strain on users. However, the study did not report system delays or performance metrics such as average frame rate and camera resolution. 

We address the above limitations in this paper by conducting a large-scale user survey with over 26 participants. In our study, participants self-reported minimal experience with robots (Mean 2.8 $\pm$ 1.2 out of 5, with 5 indicating high experience), and our approach exclusively employs a point cloud visualisation, as it has been shown to outperform other methods in creating an immersive environment \cite{su_mixed_2022}. However, comparing with the literature, such as \cite{su_mixed_2022} is not feasible due to task differences; we report our system's performance, including camera capture rate, resolution, and simulation frames per second. 
Our study uses Unreal Engine 5.4 and ROS2, making it, to our knowledge, the only teleoperation system to do so, as most studies use Unity. An exception is noted in \cite{naceri_vicarios_2021}, which used Unreal Engine 4 and ROS1. This difference may be attributed to the higher complexity of Unreal Engine and the relative scarcity of mixed reality tools \cite{coronado_integrating_2023}. %We chose to use it as we were planning to run our whole framework using a single computer.

\section{IMMERTWIN Framework}\label{sec:framework}

\begin{figure*}[t]
\centering
\vspace{3pt}
\includegraphics[width=0.9\textwidth]{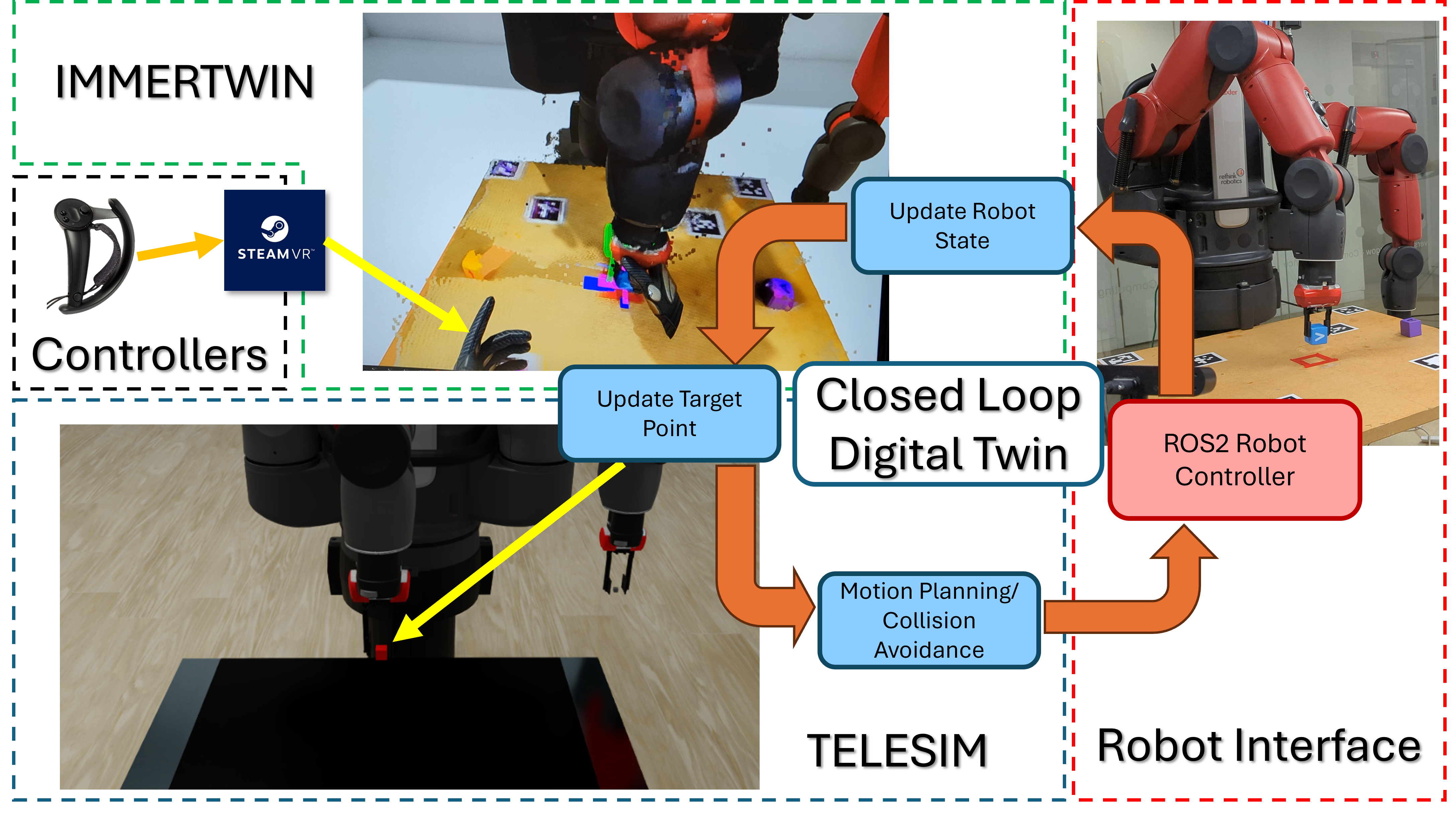}
\caption{Overview of IMMERTWIN. Our framework, IMMERWTIN, shown in the green dotted line, accepts the pose of any 3D VR controller (shown in the black dotted line) to update the position of the user's virtual hand in Unreal Engine. The user can then grab the robotic gripper and move the robot where they want. The new robotic goal is then transmitted to TELESIM (shown in the blue dotted line) to perform motion planning and collision avoidance. The state of the virtual robot is then transmitted to the real robot, shown in the red dotted line, to update its position. Finally, the state of the real robot is transmitted back into Unreal Engine to create a closed-loop digital twin. }
\label{fig:schema}
\vspace{-0.5cm}
\end{figure*}

IMMERTWIN is an add-on to TELESIM. TELESIM \cite{audonnet_telesim_2023} is a framework designed for the teleoperation of various robotic arms using any control system capable of outputting a 3D pose. For completeness, a brief overview of TELESIM is given below. TELESIM consists of three main components: The first component, \textit{the controller} which, in our previous study \cite{audonnet_telesim_2023}, consisted of a Steam Index VR controller and a Senseglove, both equipped with a Vive VR Tracker. The second component, \textit{TELESIM} itself, calculates a motion plan in real-time based on the pose provided by the controller. The third component, \textit{the robot interface}, involves transmitting the robot's joint states via ROS2 (Robot Operating System) \cite{macenski_robot_2022} to any robotic arm compatible with ROS control. TELESIM was deployed into a Rethink Robotics Baxter and a Universal Robots 3 (UR3).

IMMERTWIN functions as an enhancement to TELESIM. It modifies the TELESIM control loop by replacing the controller with a virtual gripper, depicted in bright green in Figure \ref{fig:experimental_setup} as A. This virtual gripper's position is transmitted to TELESIM, enabling Isaac Sim to compute a motion plan for the robot while ensuring collision avoidance. The robot's joint states are then relayed to the robot. In TELESIM, this would conclude its functionality; however, IMMERTWIN extends this by transmitting the real robot's state to Unreal Engine 5.4, as shown in Figure \ref{fig:schema}, which may differ from Isaac Sim's request due to unexpected collisions or hardware malfunctions. These modifications transform TELESIM, which was a digital clone to a closed loop digital twin, thus granting it the name of IMMERTWIN, allowing users to interact with the real robot via ROS2.

In IMMERTWIN's virtual environment, the virtual robot mirrors the real robot. If there are no errors, such as collisions or unreachable positions, the robot gripper aligns with the virtual gripper. Users interacting in VR can manipulate the virtual gripper to control the robot. A point cloud visualisation from two calibrated ZED 2I cameras provides feedback on the world state.  This point cloud provides users with near-real-time feedback (100ms) on the robot and its surroundings, enabling them to adjust their actions accordingly. IMMERTWIN's control loop design enables users to release the virtual gripper, maintaining the robot's position, thus allowing users to rest, reposition, or gain different perspectives, enhancing manipulation precision. 

%While new robots can be integrated into Unreal Engine, the process is more intricate than with TELESIM. TELESIM simplified the process of loading robot description files by using URDF (Universal Robot Description File), facilitating automatic loading with minimal adjustments. On the other hand, the user needs to import URDF model files into the engine. Therefore, new robots must be assembled by manually modifying the mesh and reassigning joints to their appropriate positions based on existing templates. Although the latter process is more intricate than TELESIM's, it can still be considered relatively straightforward.

We opted to use Unreal Engine on Linux as it is the only VR-capable engine on Linux. However, the Unreal Engine cannot build lighting on Linux and we needed to use LUMEN\cite{noauthor_lumen_2024}, Unreal Engine's resource-intensive real-time lighting system. Linux was required to use ROS2 and interface it with TELESIM. Furthermore, our goal was to run the entire system on a single computer to enhance portability and user-friendliness, reinforcing our plug-and-play approach. Although switching to Unreal Engine for another engine is feasible due to IMMERTWIN's modularity, this task is deferred to future work. Additionally, Unreal Engine was chosen because ROS2 integration had already been developed by Rapyuta Robotics\footnote{https://github.com/rapyuta-robotics/rclUE}. The integration's sole requirement is a manually operated ROS2 "server" node running FastDDS, functioning similarly to a ROS1 master.

\section{Experimental Setup}\label{sec:experimental_setup}

\subsection{TELESIM}\label{sec:telesim}

In our previous study \cite{audonnet_telesim_2023}, we assessed our framework through a user survey involving 37 participants (29 male and 8 female) using the following setup. We employed two robots, Baxter and UR3, each positioned in front of a table with cubes arranged in an isosceles triangular pattern, similar to the setup depicted in Figure \ref{fig:experimental_setup}. Participants were tasked with teleoperating both robots while standing with their backs to the VR headset, positioned to the left of each robot at a height of one meter. The headset served as the world's origin and a reference point for the participants. Each robot was controlled using a different method: the Baxter robot was operated with a Steam Index VR controller, as illustrated in Figure \ref{fig:experimental_setup}, while the UR3 robot was controlled using a Senseglove development kit\cite{noauthor_vr_2023}. This kit allowed for the mapping of individual finger movements, with an HTC Vive Tracker mounted on the hand. Only the thumb and index finger movements were used to control a modified T42 gripper from the Yale OpenHand project \cite{noauthor_yale_2023} mounted on the UR3. 

% Participants were first asked to complete a brief demographics questionnaire, including their age, gender, and experience with Virtual Reality, robots, and gaming wearables. These questions were posed using a 5-point Likert scale, reflecting each participant's perception and understanding. Following this, participants were briefed on the setup and task and given 5 minutes of practice time with Baxter to familiarise themselves with the system. They then had 10 minutes to stack three 40mm cubes, arranged in an isosceles triangle, to the centre of the table as many times as possible. Once the tower was built, the timer was paused, and the cubes were reset to their original positions. After teleoperating Baxter, participants completed two additional questionnaires: the Single Ease Question (SEQ) \cite{hodrien_review_2021}, a reliable end-of-task metric, and the raw NASA-TLX \cite{hart_nasa-task_2006}, which evaluates cognitive and physical load. These questionnaires were repeated after participants finished teleoperating the UR3. Finally, they completed the Negative Attitude Towards Robots Scale (NARS) \cite{syrdal_negative_2009}.

Participants completed a demographics questionnaire using a 5-point Likert scale. Following this, they were given 5 minutes to practice with TELESIM and Baxter before attempting to stack as many as possible three 40mm cubes in the centre of a table within 10 minutes; cubes were initially arranged in an isosceles triangle. Participants completed the Single Ease Question (SEQ) and the raw NASA-TLX for Baxter. After that, the same experimental methodology was repeated for the UR3. At the end of the experiment, they filled out the Negative Attitude Towards Robots Scale (NARS).

\subsection{IMMERTWIN}\label{sec:immertwin}

To ensure a fair comparison between IMMERTWIN and TELESIM, we replicated the user survey setup used for TELESIM and the robotic hardware. However, we implemented several modifications for the UR3 setup based on feedback received during the TELESIM user survey. That is, we replaced the custom-built gripper with a Robotiq 2F-85 gripper \cite{noauthor_robotiq_2024}, which has a reduced finger length, addressing issues where the previous gripper impacted robot motion due to safety settings. Additionally, the new gripper resolves issues related to insufficient grasping force, with a maximum force of 235N, preventing cubes from falling during transit. To accommodate the UR3's limited range, we mounted it on a pedestal, enhancing the robot's freedom of movement. During experiments, users reported that the UR3 felt more responsive than the Baxter robot, providing them greater control over its movements. Finally, we transitioned from using a SenseGlove to a Valve Index VR controller for both robots, as the former contributed to the UR3's suboptimal performance in TELESIM. Participants no longer need to stand close to the robots; instead, they are positioned 2 meters away, behind a physical security tape visible on the top-right in Figure \ref{fig:experimental_setup} as C. Additionally, virtual bounding boxes were added in VR to alert users when they are approaching virtual limits set to prevent them from getting too close to the robots. Users were free to move around the room by walking within their available space. For IMMERTWIN, we included the Simulator Sickness Questionnaire \cite{kennedy_simulator_1993} to assess potential symptoms such as nausea or headache caused by immersive simulation software. This questionnaire has been validated in mixed reality environments \cite{bechtel_toward_2023}. 

The cubes were arranged in an isosceles triangle, with each placement presenting varying difficulty levels, similar to \cite{audonnet_telesim_2023}. For both Baxter and the UR3, the leftmost cube was the furthest from one of the cameras, resulting in a slightly unstable image in the 3D cloud during the experiments. The other two cubes were placed at the robot's range limit, often requiring users to rotate the gripper. The tower position was at the intersection of all cubes. This position proved challenging for some Baxter participants, as the robot arm could only sometimes lift sufficiently to stack the cubes, depending on the gripper's angle. Users encountering this difficulty were advised to adjust their hand rotation. This issue was not present with the UR3.

We recruited 26 participants (21 male, 5 female), with an average age of 27.8 years and a standard deviation of 6.8, from the University of Glasgow. Initially, participants completed a brief questionnaire about their background, following the same experimental methodology as TELESIM. The experiment's conditions and objectives were then explained: participants had to manipulate a virtual gripper in a virtual world to grasp and stack three cubes arranged in an isosceles triangle in front of the robot to the centre of the table. They were given 5 minutes of practice and 10 minutes to build as many towers as possible. Users were only informed of the remaining time if they inquired. After using the first robot, Baxter, participants completed three questionnaires: the raw NASA questionnaire, the SEQ, and the Simulator Sickness Questionnaire. They then repeated the procedure with the UR3 robot.

\subsection{Hardware}

As detailed in Section \ref{sec:framework}, the point cloud data for each robot is sourced from two Zed2I cameras\footnote{https://www.stereolabs.com/en-fr/store/products/zed-2i}, positioned 120 degrees apart and oriented towards the working area. This setup encompasses the robot, the table, and the three cubes intended for user manipulation, as described in Section \ref{sec:telesim}. The point clouds are updated every 100 milliseconds, as higher refresh rates adversely affect the frame rate. The point cloud is generated from RGB and depth images transmitted to Unreal Engine using ROS2, processed via a compute shader on the GPU to enhance performance. We developed this compute shader\footnote{https://github.com/cvas-ug/immertwin} to reconstruct the 3D point cloud from RGB and Depth images transmitted by the 2 ZED2I cameras, as ROS2 point cloud messages are more expensive to parse and need to be processed on the CPU. Since compute shaders executes every frame, we limit the point cloud's refresh rate by sending images at 10Hz. The images are captured at 720p resolution since higher resolutions result in performance degradation.

To maximise performance, given VR's suboptimal optimisation on Linux, image capture and transmission were performed on an Nvidia RTX 2080TI, while the Unreal Engine ran on an Nvidia RTX 4090. Both GPUs were housed in the same computer to minimise latency that could arise from networked machines, especially when transmitting large volumes of images. Additionally, the TELESIM component of our framework was executed on a separate computer equipped with an Nvidia RTX 3060. This setup ran Isaac Sim alongside ROS2 control for robot control. The only data exchanged between the two computers via ROS2 over an Ethernet connection were the 3D pose of the virtual gripper and the real robot joint states, resulting in negligible network delay. However, the approximate 100ms delay inherent to TELESIM persisted in IMMERTWIN, primarily due to path planning and the robot's slow movement for safety reasons. This configuration allowed us to achieve an average of 40 fps on the VR headset, with minor tearing occurring if the user moved their head rapidly. Nevertheless, as described in Section \ref{sec:evaluation}, few users reported experiencing nausea or other adverse effects from using VR.

\section{Evaluation}\label{sec:evaluation}
\begin{figure}[t]
\vspace{3pt}
\includegraphics[width=0.95\linewidth]{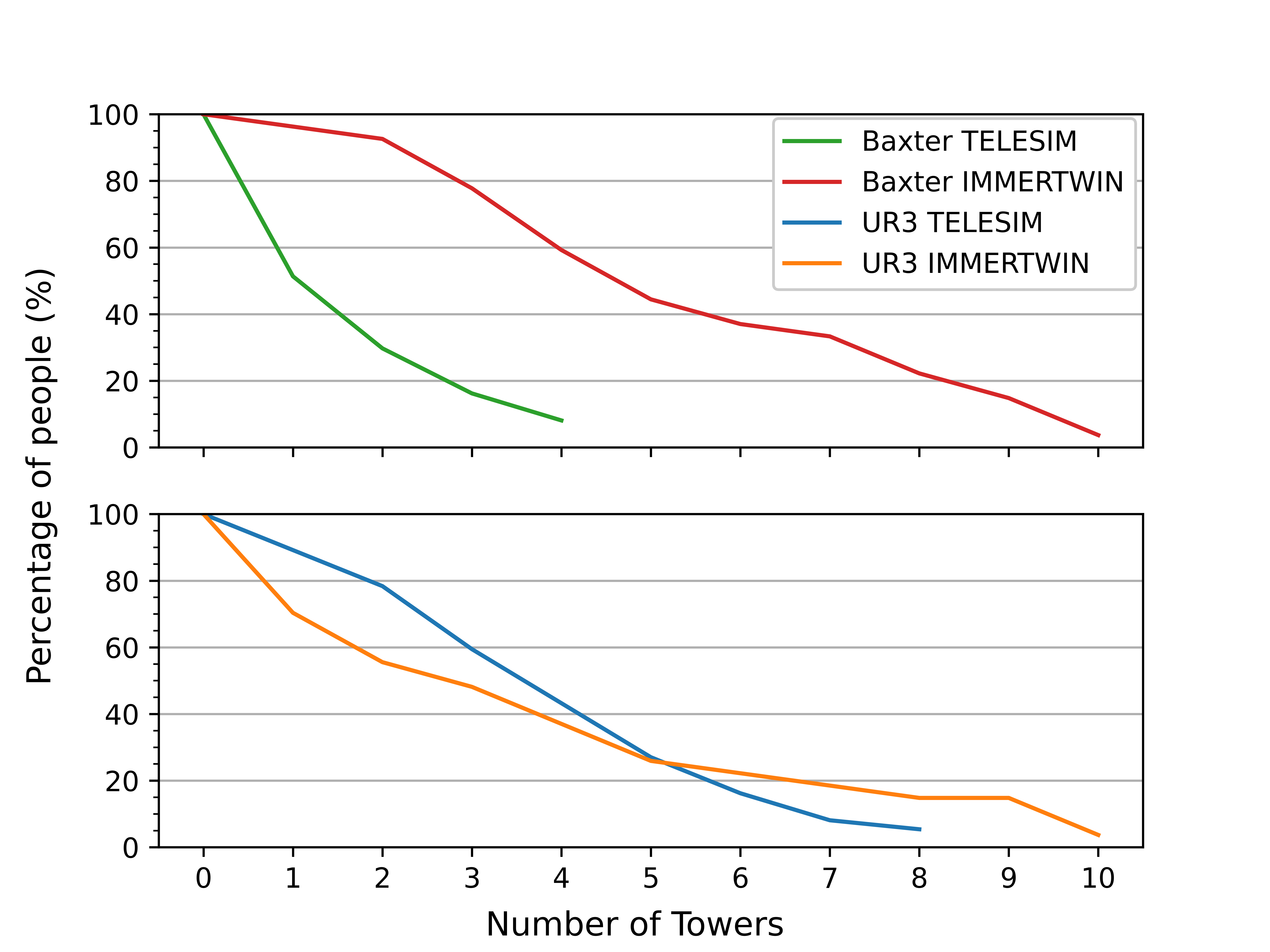}
\caption{Population percentage for each tower completed for both robots for TELESIM and IMMERTWIN, respectively.}
\label{fig:percentage}
\vspace{-0.5cm}
\end{figure}

Figure \ref{fig:percentage} shows that participants of IMMERTWIN successfully constructed a maximum of 10 towers with both robots. Notably, users operating the UR3 with IMMERTWIN consistently completed more towers than participants using any other robot across both TELESIM and IMMERTWIN. Interestingly, for the Baxter robot, participants who built fewer than 5 towers did so more frequently in TELESIM than in IMMERTWIN. This trend can be attributed to user feedback during and after the experiments, where participants noted that Baxter was less precise, particularly participants who encountered the stacking issue discussed in Section \ref{sec:immertwin}, and preferred the UR3. One reason cited was that the UR3's gripper is larger when open than Baxter's, allowing for more margin of error when grasping objects. However, participants who achieved more than 5 towers favoured Baxter over the UR3, although they did not provide specific reasons other than the UR3 being slower than Baxter. The UR3's movement speed was intentionally set low to prevent damage to the robot. This preference is also reflected in the results of the Single Ease Question, with a mean score of $3.81\pm1.52$ for the UR3 and $3.5\pm1.50$ for Baxter where a higher score indicates an easier task difficulty. Although there is no significant statistical difference, users preferred the UR3. %Furthermore, these results support the observation that more users preferred the UR3, as a greater number of participants completed fewer than 5 towers.

\begin{figure}[t]
\vspace{3pt}
\includegraphics[width=0.95\linewidth]{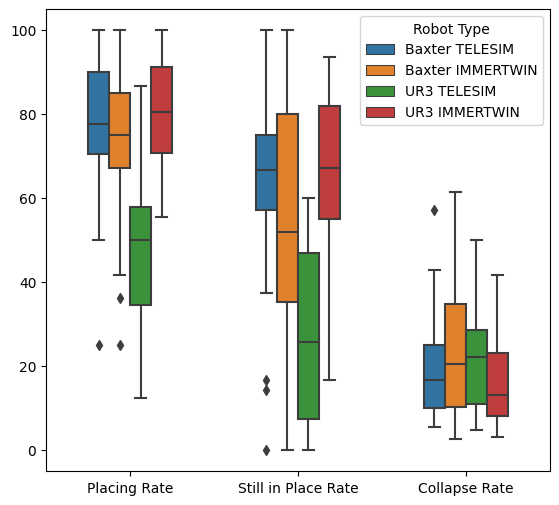}
\caption{Ratio of different statistics collected during the experiment. The Placing Rate is calculated as the number of place actions over the number of picking actions. The Collapse Rate is calculated as the number of Collapse actions over the number of picking actions. The Still in Place Rate is calculated as the number of Place actions minus the number of collapses over the number of picking actions, effectively rating the tower's stability.}
\label{fig:ratio_stats}
\vspace{-0.5cm}
\end{figure}

The perceived superiority of the UR3 for IMMERTWIN is further corroborated by Figure \ref{fig:ratio_stats}, which present the ratios of Placing Rate, Collapse Rate, and Still in Place Rate. For TELESIM, the UR3 exhibits a statistically significant difference (99.9\%) in Placing Rate, Collapse Rate, and Still in Place Rate. However, no statistical difference is observed for Baxter in TELESIM \cite{audonnet_telesim_2023} or either robot in IMMERTWIN. This indicates that IMMERTWIN does not significantly improve these statistics. Interestingly, it also suggests that the additional freedom of motion afforded by immersion in a Digital Twin does not significantly enhance teleoperation performance for certain robot types. The poor performance of the UR3 in TELESIM may be attributed to an inadequate control method or the low quality of the gripper used, as mentioned in Section \ref{sec:immertwin}. All statistical tests were conducted using the Mann-Whitney U Test. Given our relatively small sample size, a non-parametric test was deemed appropriate, as recommended by Rochon \textit{et al.} \cite{rochon_test_2012}.

\begin{figure}[t]
\vspace{3pt}
\includegraphics[width=0.95\linewidth]{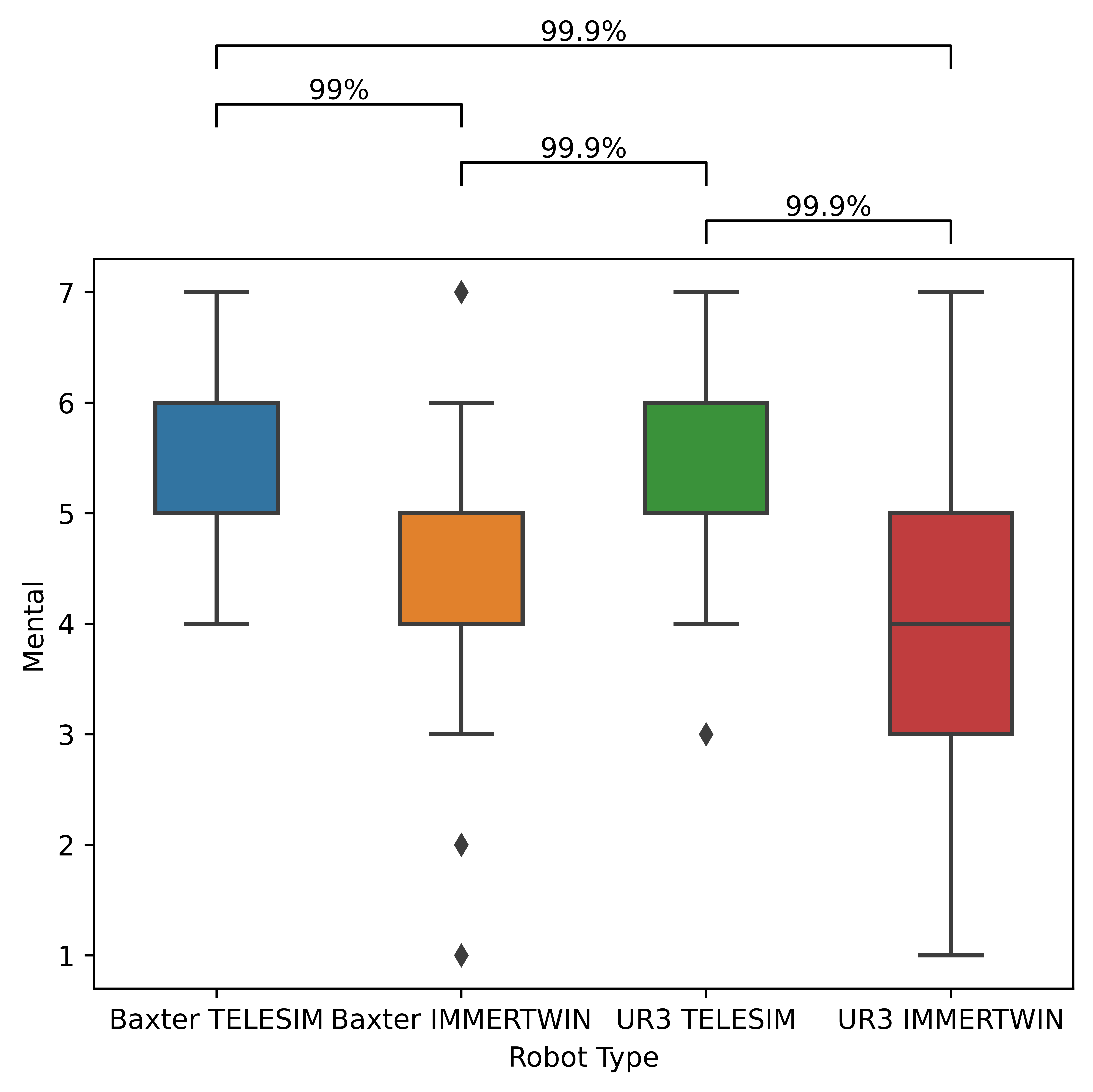}
\caption{Result of the raw NASA-TLX mental aspect, which evaluates how mentally demanding the task was. A low score indicates a lower effort. The horizontal bar at the top indicates a significance value between the 2 items indicated by the ticks at both ends of the bar.}
\label{fig:mental}
\vspace{-0.5cm}
\end{figure}

\begin{figure}[t]
\vspace{3pt}
\includegraphics[width=0.95\linewidth]{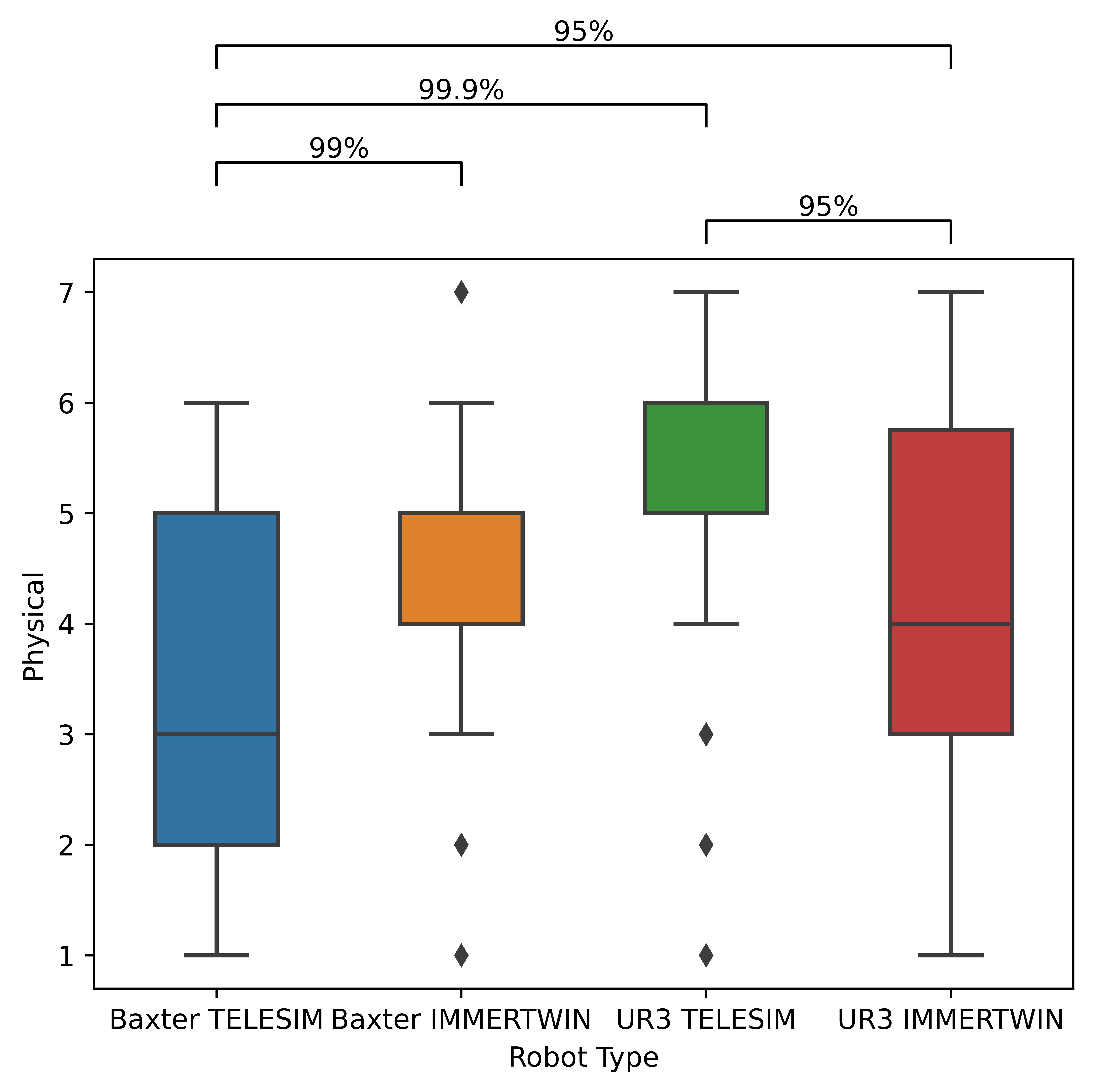}
\caption{Result of the NASA-TLX physical aspect, which evaluates how physically demanding the task was. A low score indicates a lower effort. The horizontal bar at the top indicates a significance value between the 2 items indicated by the ticks at both ends of the bar.}
\label{fig:physical}
\vspace{-0.5cm}
\end{figure}

The observation that IMMERTWIN does not yield performance gains in teleoperation does not negate the utility of VR. Among the 26 participants, 12 had previously participated in TELESIM and were asked to express their preference between the two systems. Three-quarters of these participants favoured IMMERTWIN, with only one participant preferring TELESIM. Furthermore, Figure \ref{fig:mental}, which illustrates the mental effort required during the task, indicates that IMMERTWIN demands significantly less effort than TELESIM, with at least a 99\% significance level. However, this does not hold for the physical aspect, as shown in Figure \ref{fig:physical}, where the Baxter robot in TELESIM exhibits at least a 95\% significantly lower physical effort than all other experiments. Participants using IMMERTWIN with both robots observed a similar level of exhaustion, while the UR3 in TELESIM was statistically more exhausting than the other robots. Based on user feedback, we hypothesise that IMMERTWIN causes an average level of physical exhaustion due to users needing to bend for a closer view. 

An interesting difference noted in the NASA questionnaire is the pacing. Although the experimental setup was identical for both robots, IMMERTWIN users perceived the pace as being 95\% significantly slower, with an average pace of 3.8 for IMMERTWIN compared to 4.1 for TELESIM, in which a higher number means that they felt more rushed. We attribute this difference to users losing their sense of time, as intense mental activity has been shown to cause time distortion in VR environments \cite{moinnereau_quantifying_2023}.

Finally, the Simulator Sickness Questionnaire results indicate that the most common symptoms, in order of prevalence, are fatigue, general discomfort, eyestrain, and headache. However, for most users, these symptoms are mild, with fatigue having a mean score of $1.73\pm0.92$, where 1 indicates no symptoms, and 4 indicates severe symptoms.

\section{Conclusion and Future Work}

This paper explored the capabilities and user experiences of IMMERTWIN compared to its predecessor TELESIM \cite{audonnet_telesim_2023}. That is, IMMERTWIN allows for plug-and-play teleoperation in an immersive digital twin using a VR headset, while TELESIM only allows teleoperation using direct visual feedback. Overall, IMMERTWIN demonstrates the potential of integrating VR into teleoperation frameworks, offering a more engaging and less mentally taxing user experience. IMMERTWIN is available on GitHub\footnote{\href{https://cvas-ug.github.io/immertwin}{https://cvas-ug.github.io/immertwin}}, allowing developers to perform teleoperation on their robots with minimal setup time. Despite IMMERTWIN not demonstrating significant performance improvements in teleoperation metrics, it was favoured by most participants who had previously used TELESIM. This preference underscores the potential benefits of immersive VR environments in enhancing user experience, even if quantifiable performance gains are not immediately evident. The mental effort required by IMMERTWIN was significantly lower than that of TELESIM, suggesting that the immersive nature of the VR environment may alleviate cognitive load. However, physical effort remained at the same level, particularly with the UR3 in TELESIM, which was noted to be more exhausting than other setups. This highlights the need for further refinement in balancing cognitive and physical demands in VR-based teleoperation systems. 

The study in this paper also revealed interesting insights into user perceptions of time and pacing within VR environments, with IMMERTWIN users experiencing a slower perceived pace. This aligns with existing research indicating that intense mental activity in VR can alter time perception, a factor that should be considered in future VR application designs\cite{moinnereau_quantifying_2023}. Moreover, while mild symptoms of simulator sickness were reported, they were generally manageable, indicating that IMMERTWIN's VR implementation is well within acceptable limits for user comfort. However, the long lasting effect of virtual reality teleoperation are left as future work. The feedback gathered provides valuable guidance for future iterations of IMMERTWIN, focusing on enhancing physical ergonomics and further reducing cognitive load. 

Future research consists of optimising VR systems further, exploring ways to mitigate physical strain and enhancing the intuitive control of robotic systems in immersive environments. This approach aims to leverage human intuition and a high-level task overview while allowing robots the autonomy to execute precise actions, such as accurate object picking and placing. We intend to examine whether the level of environmental realism affects teleoperation performance. Currently, the teleoperation environment is a featureless, white room with only the robot and workspace rendered. We plan to enhance this by immersing users in 3D reconstructed representation of the real environment. We hypothesize that while this change may not directly impact performance metrics, it could enhance the overall user experience. Additionally, we aim to investigate whether the increased physical exhaustion reported with IMMERTWIN compared to TELESIM is attributable to the vertical height of the virtual environment. To address this, we will provide users with the ability to adjust the environment's height to their preference, potentially reducing physical strain.Finally, we plan to replace the cube stacking task used as a benchmark for a peg-in-the-hole task which is a current manufacturing task, requiring high level of details and precision.

%%%%%%%%%%%%%%%%%%%%%%%%%%%%%%%%%%%%%%%%%%%%%%%%%%%%%%%%%%%%%%%%%%%%%%%%%%%%%%%%
%%%%%%%%%%%%%%%%%%%%%%%%%%%%%%%%%%%%%%%%%%%%%%%%%%%%%%%%%%%%%%%%%%%%%%%%%%%%%%%%

%%%%%%%%%%%%%%%%%%%%%%%%%%%%%%%%%%%%%%%%%%%%%%%%%%%%%%%%%%%%%%%%%%%%%%%%%%%%%%%%

%%%%%%%%%%%%%%%%%%%%%%%%%%%%%%%%%%%%%%%%%%%%%%%%%%%%%%%%%%%%%%%%%%%%%%%%%%%%%%%%

\bibliographystyle{IEEEtran}
\bibliography{references, lib}

\end{document}